\documentclass[11pt]{article}
\usepackage[margin=1in]{geometry}
\usepackage{newtxtext}
\usepackage{newtxmath}
\usepackage{graphicx}
\usepackage{booktabs}
\usepackage{enumitem}
\usepackage{multirow}
\usepackage{hyperref}
\usepackage{xurl}
\usepackage[section]{placeins}
\usepackage[font=small,labelfont=bf]{caption}   

\setlist[itemize]{nosep, leftmargin=*}
\setlist[enumerate]{nosep, leftmargin=*}

\pdfoutput=1

\renewenvironment{abstract}
  {%
    \begin{center}
      \textbf{ABSTRACT}
    \end{center}
    \begin{quotation}
    \normalsize
  }
  {%
    \end{quotation}
    \vspace{1.5ex}%
  }

\makeatletter
\def\verbatim@font{\small\ttfamily}
\makeatother

\title{\textbf{Think Through a Bottleneck: Hourglass Reasoning for Rigorous Induction}}
\author{Huan Zhu\thanks{Code and prompts: \url{https://github.com/ZhuHuan09/hourglass-reasoning}} \\ Peking University}
\date{}

\begin{document}

\maketitle

\begin{abstract}
Self-refinement often fails to strengthen few-shot inductive reasoning in large language models. Prompting a model to explicitly state its inferred rule does little on its own. What actually matters is a structurally enforced isolation between reasoning stages, so that information can only pass between them as a compressed symbolic state.

We introduce \textbf{Hourglass reasoning}, which enforces strict context isolation between reasoning stages. The frozen LLM acts as a meta-constructor, building for each task a symbolic encoder--decoder: an Induction module compresses the support examples into a schema $\phi$ (encoder) and a transient scaffold $z$; a Deduction module derives rule $T$ (decoder) from these and discards $z$; an Implementer compiles $(\phi, T)$ into artifacts; an error-driven Refiner revises $(\phi, T)$ and regenerates artifacts from scratch. Only $(\phi, T)$ crosses stage boundaries, so all refinement stays anchored to the rule.

We evaluate Hourglass across three benchmarks spanning visual abstraction, hardware synthesis, and textual rule induction, using GPT-5.5 and Gemini 3.1 Pro. On ARC-AGI-2, it raises best-of-5 accuracy by up to 14 points over an iterative-refinement baseline. On ChipBench, it nearly doubles Verilog synthesis accuracy with GPT-5.5, from 31\% to 58\%. BBEH-Linguini draws on puzzles from the International Linguistics Olympiad, a setting where prior work has shown that explicit verbalization can hurt performance. Hourglass mitigates this tendency, and on Gemini 3.1 Pro, it reverses the effect entirely.

Ablations confirm that these gains come from the isolation between stages and the quality of the initial induction, not from prompt wording or the particular symbolic form used. It is how information flows through the reasoning process, rather than the language used to express it, that drives inductive reasoning in frozen LLMs.
\end{abstract}

\section{Introduction}

Humans can extract abstract rules from only a handful of examples. In artificial intelligence, this capability is rigorously tested by benchmarks like ARC-AGI-2 (Chollet, 2019), where each puzzle is governed by a single, precise transformation rule that must be inferred from few-shot demonstrations. Despite impressive performance on many natural-language tasks, current frontier LLMs still struggle with such rule-centric induction.

Large language models have demonstrated remarkable capabilities across a broad spectrum, extending from natural language processing and code generation to hardware description language synthesis (Liu, Y., et al., 2025) and abstract spatial reasoning on benchmarks such as ARC-AGI-2 (Franzen et al., 2025). However, these models remain prone to shortcut learning (Geirhos et al., 2019): instead of abstracting the latent rule, a model exploits superficial regularities in the support examples. When execution feedback is available, this often manifests as patchwork logic: hardcoded if-else branches keyed to specific coordinates, example indices, or local artifacts. Such patches force support examples to pass but fail on out-of-distribution queries (Moskvichev et al., 2023; Mitchell et al., 2023). Moreover, a growing body of recent work demonstrates that the intrinsic self-correction capabilities of monolithic LLMs are brittle and frequently degrade performance in the absence of external structural guidance (Tsui, 2025; Sanz-Guerrero \& Von Der Wense, 2025).

\begin{figure}[b]
  \centering
  \includegraphics[width=0.75\linewidth]{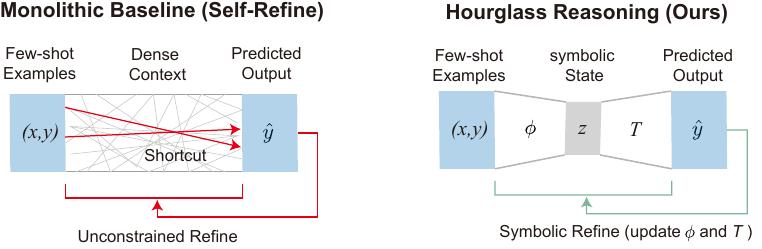}
  \caption{\small\textbf{Monolithic self-refinement vs.\ Hourglass Reasoning.}
  \textbf{Left:} Self-Refine maps few-shot examples $(x,y)$ to $\hat{y}$ through a single dense context; entangled dependencies enable \emph{shortcut} solutions (red) that fit the demonstrations without recovering the underlying rule.
  \textbf{Right:} Hourglass Reasoning compresses the examples into an explicit symbolic state, a schema $\phi$ and a transformation rule $T$, through an encoder-decoder-style reasoning topology. All intermediate reasoning, including the transient scaffold $z$, expires within isolated stage contexts; only $(\phi, T)$ is passed forward. Refinement (green) revises $(\phi, T)$ rather than the output directly.}
  \label{fig:overview}
\end{figure}

These pathologies are partly rooted in how information flows through dense, unrestricted context windows. When raw examples, current artifacts, error feedback, and repair instructions coexist without structured partition, the model tends to anchor on low-level perceptual details rather than generalizing to an abstract rule. Standard self-refinement treats refinement as artifact-level editing: it changes the current code directly, without maintaining an explicit symbolic rule as the persistent target of repair (Madaan et al., 2023). Simply prompting the model to ``think step by step'' or ``explain the rule first`` within the same context offers only a soft constraint: when feedback arrives, the model can abandon its earlier rule and patch the code directly (Huang et al., 2023). Figure~\ref{fig:overview} contrasts this monolithic failure mode with the alternative we propose: routing refinement through an explicit symbolic bottleneck rather than editing the output directly.

To address this, we impose a structured bottleneck on inductive reasoning: \textbf{Hourglass reasoning}.The LLM acts as a meta-constructor that, for each task, dynamically builds a symbolic encoder--decoder. A schema $\phi$ serves as the encoder that parses inputs, and a transformation rule $T$ serves as the decoder that generates outputs. Crucially, context isolation between stages ensures that only the compressed symbolic state $(\phi, T)$ passes forward, while all intermediate reasoning traces—including a transient scaffold $z$—are discarded. This forces all downstream implementation and refinement to operate exclusively through the rule, preventing instance-specific details from leaking across stages. The full architecture is presented in Section~3.

We evaluate Hourglass on three benchmarks spanning visual abstraction, formal hardware synthesis, and textual rule induction (rationale detailed in \S 4.1). Across all three, the same domain-agnostic pipeline yields substantial and consistent gains over a context-reset Self-Refine baseline. Ablation studies on ARC-AGI-2 reveal a striking robustness: so long as the core workflow topology is preserved, the gains remain stable even when all auxiliary prompt directives are stripped away. Structured self-refinement that outputs identical symbolic intermediates but lacks context isolation performs substantially worse, confirming that the topology itself, not prompt engineering or output formatting, is the key contributing factor.

Our contributions are:

\begin{enumerate}
\item \textbf{A structured bottleneck for inductive reasoning}, realized through role-isolated symbolic state passing. Ablations show that isolation between stages is necessary for the gains to hold, while the specific symbolic vocabulary used to express $(\phi, T)$ can vary substantially without much loss in performance, indicating that how information is allowed to flow through the reasoning process, more than the particular form it takes, is what drives the improvement.

\item \textbf{Cross-domain empirical evidence.} The same domain-agnostic pipeline yields consistent, substantial gains on three unsaturated benchmarks spanning visual abstraction, formal hardware synthesis, and anti-prior linguistic deduction.

\item \textbf{Ablations that isolate the key contributing factor.} We show that structured self-refinement alone fails to reproduce the gains, while a minimal role-isolated variant preserving only the $(\phi, z, T)$ topology remains equally strong, demonstrating that the workflow topology---not prompt engineering---drives the improvement.
\end{enumerate}

\section{Related Work}

\subsection{Test-time computation and self-refinement}

Allocating additional test-time compute via sampling, search, or verification can improve LLM outputs (Snell et al., 2024). Aggregation methods such as self-consistency reduce variance by marginalizing over reasoning paths (Wang et al., 2023). However, unconstrained self-refinement can amplify local mistakes when each iteration edits the current artifact without an explicit abstract state (Madaan et al., 2023; Huang et al., 2023). In fact, several recent studies report that naive self-refinement can even degrade performance due to context contamination and premature convergence (Sanz-Guerrero et al., 2025; Tsui et al., 2025). The intuition that discarding instance-specific detail can improve generalization is closely related to the information-bottleneck perspective we adopt below (\S 2.3); Hourglass translates this intuition into a practical symbolic-state workflow rather than a mathematically enforced compression.

A separate line of work pursues a complementary strategy: instead of refining within a single episode, these methods accumulate reusable problem-solving strategies across episodes at test time, without updating model weights, such as a persistent, self-curated memory of strategies and code snippets (Suzgun et al., 2026) or a library of procedural skills distilled from past trajectories (Yang et al., 2026). These approaches typically accumulate unstructured or loosely structured natural-language strategies within a single model; we return to this contrast in \S 6.3.

\subsection{Multi-agent systems and context isolation}

Multi-agent frameworks improve performance by decomposing reasoning across specialized roles (Hong et al., 2023; Qian et al., 2023). Yet role separation alone does not guarantee context isolation (Du et al., 2023; Yin et al., 2023): when agents communicate via high-dimensional natural-language traces, irrelevant or stale signals can propagate. Hourglass adopts a stricter approach: each role is an isolated API call with no shared conversational history, and only selected symbolic artifacts, $\phi$ and $T$, are explicitly passed forward, while all intermediate reasoning traces and transient scaffolds are compartmentalized. This demonstrates that explicit, typed context isolation within a single-model pipeline can capture much of the benefit attributed to multi-agent dynamics while reducing coordination overhead and contamination risk.

\subsection{Information Bottleneck Theory}

The Information Bottleneck (IB) principle posits that optimal representations retain task-relevant information while discarding instance-specific details (Tishby et al., 1999). This idea has been concretely realised in several classical architectures. Denoising autoencoders explicitly corrupt inputs during training, forcing models to recover robust features instead of exploiting superficial statistics (Vincent et al., 2008). Variational autoencoders (VAEs) further impose a Gaussian prior on the latent space, regularising representations into a continuous, structured manifold (Kingma \& Welling, 2014). In tasks requiring strong spatial or geometric regularity, such as continuous control, this distributional constraint is often sharpened into a topological one: latent variables are restricted to the unit hypersphere via von Mises-Fisher priors, eliminating scale degrees of freedom and markedly improving out-of-distribution generalisation (Davidson et al., 2018). 

These varied instantiations share a common insight: deliberately restricting the expressivity or geometry of intermediate representations severs the propagation of irrelevant variation, thereby strengthening generalisation.

\subsection{Program synthesis and neuro-symbolic abstraction}

Neuro-symbolic approaches have long leveraged explicit symbolic intermediates to improve generalization in program synthesis (Wong et al., 2021). Systems like DreamCoder learn libraries of reusable functions that serve as a discrete bottleneck, forcing the model to re-express solutions in terms of abstract primitives (Ellis et al., 2021). Hourglass follows a similar intuition but operates at the level of natural language: instead of learning a formal DSL, it prompts the LLM to generate an ad-hoc symbolic schema ($\phi$) and transformation rule ($T$) for each task, then uses these as the sole persistent state for refinement and regeneration. This design retains the benefits of a discrete intermediate without requiring a pre-defined grammar or training phase, thus remaining applicable across a wide range of domains.

\section{Methodology: Hourglass Reasoning}

\begin{figure}[t]
  \centering
  \includegraphics[width=\linewidth]{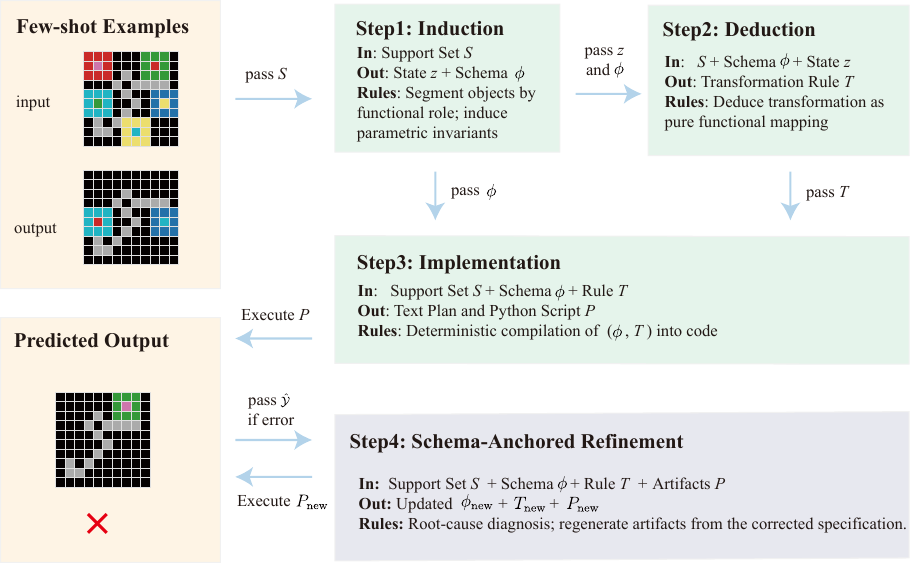}
  \caption{\small\textbf{The four-stage Hourglass Reasoning pipeline (illustrated for ARC-AGI-2).}
  green boxes (Steps~1--3) are the one-shot initialization stage; purple box (Step~4) is the iterative refinement loop; yellow boxes represent data (support set $S$ and predicted output $\hat{y}$).
  Step~1 (Induction) produces a schema $\phi$ and a transient scaffold $z$.
  Step~2 (Deduction) derives the transformation rule $T$ from $S$, $\phi$, and $z$, then discards $z$.
  Step~3 (Implementation) deterministically compiles $(\phi, T)$ into executable artifacts $P$.
  If errors occur, Step~4 (Schema-Anchored Refinement) revises $(\phi, T)$ and regenerates $P$ in a fresh context; corrections are anchored to the symbolic state rather than patched onto the output.}
  \label{fig:pipeline}
\end{figure}

Figure~\ref{fig:pipeline} gives an overview of the resulting pipeline; we describe each stage in turn below.

\subsection{Design Rationale}

Hourglass distills the information-bottleneck principle into a structural partition of the inductive reasoning process. The central idea is to isolate the extraction of an input–output rule from its subsequent execution and debugging, with a symbolic interface serving as the sole channel between stages. By forcing the model to commit to an explicit schema $\phi$ (how to parse inputs) and a declarative transformation rule $T$ (how to map parsed structures to outputs), and by discarding all transient reasoning traces, we create a discrete bottleneck. This ensures that implementation and refinement operate exclusively on abstract, domain-level specifications rather than on raw perceptual details or ad-hoc patches. Each stage is a stateless, isolated API call; only $(\phi, T)$, the support set $S$, and the current artifacts are forwarded, while all intermediate scaffolds and error histories are contained within their respective sessions. The following subsection enumerates the concrete components that realize this design.

\subsection{Components and Information Flow}

The persistent state carried across iterations is the pair $(\phi, T)$. The scaffold $z$ is transient: it aids the derivation of $T$ but is purged before any downstream processing. All three are expressed in lightly structured natural language. For an ARC-AGI-2 task, $\phi$ might capture rules for extracting connected components and their relative bounding boxes; $z$ would enumerate the specific components found in each support grid; $T$ would specify the geometric transformation applied to each component.

From this symbolic specification, the pipeline produces executable artifacts $P$. Depending on the benchmark, $P$ consists of either code-and-text or text-only depending on the task. $P$ is regenerated from scratch whenever $(\phi, T)$ is updated, but is not itself part of the persistent symbolic state.

Hourglass is realized through four stages, each an isolated API call with no shared conversational history. The support set $S$ is supplied to every call as a stable factual ground; all other intermediate reasoning traces---draft scaffolds, justifications, rejected code versions, and prior error logs---are isolated within their respective sessions.

\begin{table}[htbp]
\centering
\small
\begin{tabular}{@{}p{2.2cm}p{2.6cm}p{1.8cm}p{7cm}@{}}
\toprule
\textbf{Stage} & \textbf{Input} & \textbf{Output} & \textbf{Role} \\
\midrule
Induction & $S$ & $\phi, z$ & \textbf{Generate} a parsing schema $\phi$ and a transient scaffold $z$ from the support examples. \\
Deduction & $S, \phi, z$ & $T$ & \textbf{Derive} a transformation rule $T$ from $\phi$ and $z$; $z$ is permanently discarded after this step. \\
Implementation & $S, \phi, T$ & $P$ & \textbf{Compile} $(\phi, T)$ into executable code-and-text or text-only artifacts. \\
Refinement & $S, \phi, T, P, \hat{y}$ & $\phi_{\text{new}}, T_{\text{new}},$ \newline $P_{\text{new}}$ & \textbf{Revise} $(\phi, T)$ by comparing predicted outputs $\hat{y}$ against ground-truth $y$ on the support set; regenerate $P$. \\
\bottomrule
\end{tabular}
\caption{Stages, inputs, outputs, and roles in Hourglass.}
\label{tab:stages}
\end{table}

In addition to the context isolation, one auxiliary design choice supports interpretability: Induction and Deduction use structured output templates with labeled fields and origin-independence triggers (e.g., ``use relative coordinates''), keeping $\phi$ and $T$ explicit and auditable across iterations. This is not a mechanism we claim to be load-bearing in itself; \S 5 examines its actual contribution.

\subsection{Iterative Execution--Feedback Loop}

Starting from Induction and Deduction, after $z_0$ is discarded, the Implementer produces initial artifacts $P_0$. A maximum of $M$ refinement iterations are allowed. In each iteration, the Executor runs the current artifacts on the support inputs and produces predicted outputs $\hat{y}_t$. These predictions are passed to the Refiner alongside the ground-truth outputs $y$ of the support set, together with the current symbolic state and artifacts: $(S, \phi_t, T_t, P_t, \hat{y}_t, y)$. The Refiner produces an updated symbolic state $(\phi_{t+1}, T_{t+1})$ and regenerated artifacts $P_{t+1}$ in a fresh context. The loop halts as soon as all support examples are perfectly matched ($\hat{y}_t = y$). For a novel query $x_q$, the final compiled artifact instantiates a fresh per-input state by applying $\phi$ to $x_q$, then applies $T$ to produce the output.

In the current instantiation, the symbolic state $(\phi, T)$ is discarded once a task is solved and re-derived from scratch for the next; \S 6.3 discusses the possibility of extending it into a persistent, cross-task form.

\section{Experiments}

\subsection{Common Setup}

\textbf{Models and infrastructure.} All experiments use two proprietary LLMs: \textbf{GPT-5.5} and \textbf{Gemini 3.1 Pro}. GPT-5.5 uses medium reasoning effort; Gemini 3.1 Pro uses high reasoning effort. Each module that produces syntactically invalid output is retried up to three times.

\textbf{Baseline.} The primary baseline is Self-Refine (Madaan et al., 2023), a widely adopted iterative refinement method. In each refinement step, the model receives the support set, the current artifact, and the latest validation feedback within a single context, and then outputs a revised artifact. Both Hourglass and Self-Refine use at most five refinement iterations. No earlier rejected drafts or hidden reasoning traces are retained.

\textbf{Evaluation protocol.} For binary-correctness benchmarks, pass@1 is the fraction of tasks solved on the first mandatory run, and pass@5 is the fraction solved by any of up to five independent runs. On ARC-AGI-2, we follow the official public evaluation protocol and report the averaged score; full details of the competition-mode configuration are provided in Appendix A. Test ground truth is used exclusively in the final offline evaluation. Cost estimates are based on official API pricing, using token counts logged during runs.

\textbf{Benchmark selection rationale.} We deliberately selected three benchmarks spanning a gradient of inductive reasoning demands: \textbf{ARC-AGI-2} (Chollet, 2019), requiring pure visual rule discovery from minimal examples with no prior domain knowledge; \textbf{ChipBench} (Yu et al., 2026), a zero-tolerance benchmark for hardware logic synthesis under dynamically generated test vectors; and \textbf{BBEH-Linguini}, the Linguistics Olympiad subset of BIG-Bench Extra Hard (Kazemi et al., 2025), a setting where prior work has shown that forcing LLMs to explicitly verbalize induced rules can degrade performance (Goyal \& Dan, 2025). Full task descriptions are given in \S 4.2--\S 4.4 respectively. Together, these three benchmarks probe complementary facets of inductive reasoning: sparse visual rule discovery, knowledge-rich formal synthesis, and prior-sensitive textual induction.

\subsection{ARC-AGI-2}

ARC-AGI-2 evaluates few-shot spatial induction: each puzzle presents input--output grid pairs governed by a single, precise transformation rule, and the solver must infer this rule from the demonstrations alone. We use the public 120-puzzle evaluation set and follow the official protocol, where each task contains multiple test cases. A solution is scored by the fraction of test cases perfectly matched (pixel-for-pixel), and the overall score is the average across all tasks.

\textbf{Setup.} The primary baseline is a context-reset Self-Refine loop aligned with Hourglass on task definitions, execution feedback format, and refinement limit. It directly outputs executable code without any structured intermediate representation. To isolate the effect of output format, a Structured Self-Refine ablation (\S 5) is also reported.

\textbf{Results.} Table~\ref{tab:arc} reports the official averaged scores.

\begin{table}[htbp]
\centering
\small
\begin{tabular}{@{}lccc@{}}
\toprule
\textbf{Model} & \textbf{Self-Refine} & \textbf{Hourglass} & $\Delta$ \textbf{(Hourglass $-$ Self-Refine)} \\
\midrule
GPT-5.5 & 51.9 / 62.8 & 60.6 / 76.8 & +8.7 / +14.0 \\
Gemini 3.1 Pro & 54.4 / 76.9 & 62.4 / 86.7 & +8.0 / +9.8 \\
\bottomrule
\end{tabular}
\caption{ARC-AGI-2 score (pass@1 / pass@5, \%).}
\label{tab:arc}
\end{table}

\textbf{Analysis.} Hourglass substantially outperforms Self-Refine on both models. A plain variant that removes all auxiliary prompts while preserving the isolated $(\phi, z, T)$ topology achieves 79.3\% best-of-5 on GPT-5.5 and 87.5\% on Gemini (\S 5), indicating that role-isolated symbolic state passing, not prompt wording, is the key contributor.

A competition-mode configuration exploiting the official two-submission protocol further improves the cost--accuracy frontier, reaching 88\% accuracy at approximately \$4 per task, with an oracle variant attaining 95\% at roughly \$11 per task. Full details and comparisons against leaderboard baselines are provided in Appendix A.

\subsection{ChipBench: Hardware Logic Synthesis}

ChipBench tests hardware reasoning under zero-tolerance verification. The benchmark provides natural-language design specifications that vary in their few-shot support: some include full input--output examples, others only fragments or natural language descriptions. We evaluate both core sub-tasks: \textbf{Reference Model Generation} (Python) and \textbf{Verilog Synthesis} (synthesizable RTL). The 45 specifications span self-contained modules, CPU-related IP blocks, and hierarchical designs. A solution is correct only if it passes 100 dynamically generated random test vectors.

\textbf{Setup.} To create a uniform information condition, a preliminary LLM call reorganizes the raw specification into a \textbf{Symbolic Codebook} containing behavior tables, state transitions, priority rules, and bit-width constants. The \textbf{Native Baseline} (the official baseline provided by the ChipBench project) operates directly on the raw specification, while Codebook-SR and Hourglass each independently apply the same codebook-construction prompt before downstream reasoning. The resulting codebooks are structurally similar but independently sampled, ensuring that performance differences stem from the reasoning workflow rather than asymmetric preprocessing.

\textbf{Results.} Tables~\ref{tab:chip-ref} and~\ref{tab:chip-verilog} summarize the pass@1 and pass@5 scores.

\begin{table}[htbp]
\centering
\small
\begin{tabular}{@{}lcccc@{}}
\toprule
\textbf{Model} & \textbf{Native Baseline} & \textbf{Codebook-SR} & \textbf{Hourglass} & $\Delta$ \textbf{best baseline} \\
\midrule
GPT-5.5 & 60.0 / 66.7 & 55.6 / 64.4 & 73.3 / 82.2 & +13.3 / +15.5 \\
Gemini 3.1 Pro & 53.3 / 62.2 & 55.6 / 57.8 & 75.6 / 82.2 & +20.0 / +20.0 \\
\bottomrule
\end{tabular}
\caption{ChipBench Reference Model Generation (pass@1 / pass@5, \%).}
\label{tab:chip-ref}
\end{table}

\begin{table}[htbp]
\centering
\small
\begin{tabular}{@{}lcccc@{}}
\toprule
\textbf{Model} & \textbf{Native Baseline} & \textbf{Codebook-SR} & \textbf{Hourglass} & $\Delta$ \textbf{best baseline} \\
\midrule
GPT-5.5 & 31.1 / 35.6 & 44.4 / 51.1 & 57.8 / 66.7 & +13.4 / +15.6 \\
Gemini 3.1 Pro & 40.0 / 44.4 & 51.1 / 57.8 & 53.3 / 62.2 & +2.2 / +4.4 \\
\bottomrule
\end{tabular}
\caption{ChipBench Verilog Synthesis (pass@1 / pass@5, \%).}
\label{tab:chip-verilog}
\end{table}

\textbf{Analysis.} Hourglass outperforms the strongest baseline in all four settings. The codebook alone offers only limited benefit; the architectural bottleneck is the decisive factor. The gain of over 30 percentage points on Verilog synthesis with GPT-5.5 is particularly notable on this non-saturated benchmark. It suggests that the symbolic intermediate descriptions act as a compliance scaffold, enforcing exhaustive and consistent instantiation of known rules far more effectively than end-to-end code correction.

\subsection{BBEH-Linguini: Textual Rule Induction}

BBEH-Linguini, the Linguini subset of BIG-Bench Extra Hard (Kazemi et al., 2025), consists of puzzles drawn from the International Linguistics Olympiad. Each task provides a small set of language pairs; the solver must induce the underlying transformation rule and apply it to held-out queries. Executable code is explicitly disallowed---the model must generate and follow purely textual rules. Solving these puzzles effectively draws on structural knowledge of linguistic categories (e.g., gender, number, case). Recent work has identified a structural weakness in LLMs' explicit reasoning on such tasks (Lian et al., 2025, Choudhary et al., 2025), rendering BBEH a stress-test for pure symbolic abstraction.

\textbf{Setup.} The original problems are presented in diverse, unstandardized formats. To enable automated execution feedback via exact string matching, we used an LLM to extract each problem's demonstration pairs and test questions into a standardized JSON structure. All extractions were manually verified for character-level fidelity (full protocol in Appendix C), yielding a clean set of 144 tasks. The Raw Prompt baseline uses the original problem stem directly, without any intermediate rule extraction or refinement. During refinement, feedback is computed by exact-string comparison between predicted outputs and ground-truth answers on the support items; test answers are used exclusively for final pass@k scoring. Because code execution is unavailable, the Executor for Self-Refine and Hourglass is replaced by a separate \textbf{Inferer} agent that takes the induced rule and applies it to test inputs in a purely textual manner, analogous to an interpreter executing code (details in Appendix C.2). The Implementer produces a textual derivation plan instead of executable code; the rest of the Hourglass topology is unchanged.

\textbf{Results.} Table~\ref{tab:bbeh} summarizes the findings.

\begin{table}[htbp]
\centering
\small
\begin{tabular}{@{}lcccc@{}}
\toprule
\textbf{Model} & \textbf{Raw Prompt} & \textbf{Self-Refine} & \textbf{Hourglass} & $\Delta$ \textbf{(Hourglass $-$ Raw Prompt)} \\
\midrule
GPT-5.5 & 58.3 / 67.4 & 25.0 / 27.8 & 46.5 / 63.1 & $-$11.8 / $-$4.3 \\
Gemini 3.1 Pro & 64.6 / 68.1 & 32.6 / 33.3 & 63.1 / 79.9 & $-$1.5 / +11.8 \\
\bottomrule
\end{tabular}
\caption{BBEH-Linguini results (pass@1 / pass@5, \%).}
\label{tab:bbeh}
\end{table}

\textbf{Analysis.} Consistent with prior observations, LLMs struggle with explicit rule induction in this setting. Self-Refine severely degrades performance on both models, often falling below even the Raw Prompt. This collapse can be attributed to two intertwined mechanisms. First, \textbf{lossy compression}: forcing an LLM to verbalize its inductive insight as a natural-language rule discards subtle but critical details, and each additional refinement step compounds the information loss (Gu et al., 2024). Second, \textbf{context entanglement}: in the monolithic Self-Refine loop, rule generation, execution, and error feedback coexist in a single context, making the model prone to attention distraction (Shi et al., 2023; Liu, N. F., et al., 2024) and inconsistent rule application (Dahl et al., 2024; Zhang et al., 2025).

Unlike Self-Refine, Hourglass roughly recovers the performance of the Raw Prompt on GPT-5.5 and surpasses it by 11.8\% at pass@5 on Gemini 3.1 Pro. This suggests that the structured bottleneck may mitigate the lossy compression and context entanglement inherent in explicit rule induction, yielding rules that are sufficiently faithful for a generic Inferer to execute reliably.

\section{Analysis}

Hourglass introduces three architectural components beyond the monolithic Self-Refine baseline: a compression stage (Induction, producing $\phi$ and $z$), a reconstruction stage (Deduction, deriving $T$ and discarding $z$), and an isolated Refiner that revises $(\phi, T)$ rather than editing code directly. The prompts driving these stages each contain three classes of information (see Appendix D): a task description, auxiliary guiding prompts, and constraints on the output format of $\phi$ and $T$. The observed gains could therefore originate from the auxiliary prompts, from the hand-designed output format, from the mere presence of explicit symbolic intermediates, or from the physically isolated workflow topology itself.

To isolate the key contributing factor, we conduct five ablations on ARC-AGI-2, each disabling one candidate mechanism while preserving the others (Table~\ref{tab:ablation}).

\paragraph{Weak Initialization (Weak-Init).} \textbf{\textit{Purpose.}} If the Refiner independently drove the gains, substituting a weak Induction/Deduction model should still recover strong performance; if not, the initial bottleneck quality limits what refinement can achieve. \textbf{\textit{Setup.}} The Induction, Deduction, and Implementation modules are replaced with Gemini 2.0 Flash, while the Refiner continues to use the strong backbone.

\paragraph{Code-Only Refinement (Code-Only).} \textbf{\textit{Purpose.}} If $(\phi, T)$ is the indispensable carrier of the bottleneck, removing it from the Refiner should substantially degrade performance. \textbf{\textit{Setup.}} The Refiner no longer receives $\phi$ and $T$; it sees only the current failing artifact, the support set, and validation feedback.

\paragraph{Hourglass-Plain (Plain).} \textbf{\textit{Purpose.}} If the gains originate from careful prompt engineering, removing all auxiliary instructions should impair performance. \textbf{\textit{Setup.}} All auxiliary prompts (e.g., ``ensure origin-independence'') are stripped; only the minimal task description and output format constraints remain.

\paragraph{Hourglass-Unstructured (Unstructured).} \textbf{\textit{Purpose.}} If the hand-designed output format of $\phi$ and $T$ is crucial, removing all formatting constraints should degrade accuracy. \textbf{\textit{Setup.}} All constraints on the internal structure of $\phi$ and $T$ are lifted.

\paragraph{Structured Self-Refine (Struct-SR).} \textbf{\textit{Purpose.}} If producing structured intermediate descriptions alone is sufficient, a monolithic Self-Refine loop that outputs the same $(\phi, T)$ as Hourglass should approach its performance; failure would indicate that physical context isolation is a necessary condition. \textbf{\textit{Setup.}} The Self-Refine baseline is modified to output $\phi$ and $T$ in the same format, but all generation, execution, and feedback remain inside a single undifferentiated context window.

\begin{table}[htbp]
\centering
\small
\begin{tabular}{@{}lccccc@{}}
\toprule
\textbf{Variant} & \textbf{GPT-5.5} & \textbf{Gemini 3.1 Pro} & \multicolumn{2}{c}{$\Delta$ \textbf{vs. Full}} \\
& p@1/p@5 & p@1/p@5 & (GPT-5.5) & (Gemini) \\
\midrule
Full & 60.6 / 76.8 & 62.4 / 86.7 & --- & --- \\
Weak-Init & 40.1 / 61.4 & 54.3 / 74.6 & $-$20.5 / $-$15.4 & $-$8.1 / $-$12.1 \\
Code-Only & 58.1 / 74.7 & 72.8 / 79.9 & $-$2.5 / $-$2.1 & +10.4 / $-$6.8 \\
Plain & 59.9 / 79.3 & 70.1 / 87.5 & $-$0.7 / +2.5 & +7.7 / +0.8 \\
Unstructured & 58.6 / 76.4 & 73.6 / 89.0 & $-$2.0 / $-$0.4 & +11.2 / +2.3 \\
Struct-SR & 41.5 / 59.9 & 31.6 / 62.8 & $-$19.1 / $-$16.9 & $-$30.8 / $-$23.9 \\
SR (baseline) & 51.9 / 62.8 & 54.4 / 76.9 & $-$8.7 / $-$14.0 & $-$8.0 / $-$9.8 \\
\bottomrule
\end{tabular}
\caption{Ablation results on ARC-AGI-2 (pass@1 / pass@5, \%). Negative $\Delta$ values indicate a performance drop relative to Full Hourglass.}
\label{tab:ablation}
\end{table}

\textbf{Interpretation.} The ablations converge on a clear narrative. Auxiliary prompts, output format constraints, and even the explicit textual form of $\phi$ and $T$ are \textbf{not} the source of the gains. Plain matches Full despite removing all auxiliary instructions; Unstructured matches Full despite removing formatting constraints; Code-Only incurs only a marginal drop despite removing $\phi$ and $T$ from the Refiner entirely. In each case the gain persists so long as the role-isolated topology is preserved, ruling out prompt engineering and template design as the enabling mechanism. Notably, the Code-Only variant, which removes $\phi$ and $T$ from the Refiner's input, retains competitive performance, suggesting that the code, having been compiled from $(\phi, T)$, may implicitly retain the structure of the symbolic state.

In contrast, \textbf{physical context isolation and initial bottleneck quality are both necessary conditions}. When isolation is removed (Struct-SR), structured intermediates actively impair performance, collapsing below the unconstrained SR baseline. When the initial compression is degraded (Weak-Init), the Refiner cannot compensate, and accuracy falls sharply. Neither a strong Refiner nor structured outputs alone suffice; the bottleneck must be both physically enforced and initially competent.

Taken together, these results identify the \textbf{role-isolated symbolic topology} as the key contributing factor. In this topology, induction, deduction, and execution are separated across fresh contexts, and the transient scaffold is discarded after use. The topology is robust to substantial variation in prompting and output format, yet critically depends on the two conditions identified above.

On Gemini 3.1 Pro, Full Hourglass yields a lower pass@1 than Plain, Unstructured, and Code-Only, but recovers or surpasses them at pass@5. This pattern may reflect a trade-off: a stricter bottleneck may lock in early errors, reducing single-trial success, while the resulting abstraction quality pays off across multiple samples.

\section{Discussion}

\subsection{Core Findings}

Hourglass demonstrates that a simple structural reorganization of inference yields consistent and substantial gains across visual, hardware synthesis, and linguistic reasoning domains. This reorganization enforces a bottleneck: information must pass through it as a compressed symbolic state. Ablation evidence (\S 5) identifies the role-isolated topology, rather than prompt wording or output format, as the key contributing factor, contingent on two necessary conditions: physical isolation between stages and a sufficiently competent initial induction.

Beyond these performance gains, the explicit symbolic intermediates $(\phi, T)$ offer transparency into the model's abstraction process, providing an interpretable reasoning trace with independent value for debugging and analysis. The domain-agnostic nature of Hourglass, its avoidance of task-specific priors, and its efficiency align with the original design intent of ARC-AGI as a benchmark of general fluid intelligence.

\subsection{Limitations}

\begin{enumerate}
\item \textbf{Soft bottleneck.} $\phi$ and $T$ are natural language descriptions whose separation from implementation is enforced through prompting and context isolation rather than architectural constraints. This prompt-level design makes the method plug-and-play across frozen models, but it also means the bottleneck carries no formal guarantee.

\item \textbf{Task scope.} The current instantiation is designed for tasks with precise, deterministic rules; extending the approach to probabilistic, ambiguous, or context-dependent regularities remains an open direction.

\item \textbf{Computational overhead.} Role isolation requires disjoint API calls, increasing token usage and wall-clock latency. On GPT-5.5, a single run averages 61,911 tokens across 3.7 calls, compared to 19,594 tokens over 1.4 calls for the monolithic Self-Refine baseline---roughly a $3\times$ increase; on Gemini, 122,007 tokens (4.4 calls) vs. 61,426 tokens (1.7 calls), approximately a $2\times$ increase. Our main comparisons are therefore not compute-matched.

\item \textbf{Model coverage and evaluation protocol.} Our experiments are limited to two model families, and pass@5 estimates use early stopping, which may introduce a slight optimistic bias relative to a fully budgeted sweep. We did not conduct budget-matched runs due to resource constraints, though the magnitude of the observed gains suggests this is unlikely to be the dominant factor.

\item \textbf{Public evaluation set.} ARC-AGI-2 uses a public evaluation set; the possibility of pretraining exposure remains a shared limitation of all methods evaluated on this benchmark.
\end{enumerate}

\subsection{Future Directions}

In its current form, Hourglass is an ephemeral, per-task reasoning scaffold, strictly limited to precise, deterministic logic. Each task requires the system to induce its rules from scratch, and the scaffolding is discarded once solved.

Humans, however, can learn not just case-specific solutions but transferable meta-strategies that travel across domains. Consider a meteorologist faced with a novel time series of atmospheric pressure readings. Without task-specific training, she draws on the same mental routine: look for periodicities, segment the signal into regimes, identify anomalies. Later, given an unfamiliar economic indicator, she deploys that routine again. These meta-strategies do not prescribe exact operations, nor do they guarantee correctness, yet they systematically elevate her analysis above random guessing.

Can a model be built to learn such meta-strategies? Recent RLVR-trained models acquire reasoning primitives for iterative hypothesis generation and self-correction, but only implicitly, as opaque byproducts of large-scale training rather than explicit, reusable templates (OpenAI, 2024; DeepSeek-AI, 2025; Wen et al., 2026). Test-time accumulation methods maintain explicit strategies, but typically as unstructured memory stores matched by similarity, which limits generalization to novel abstract rules (Suzgun et al., 2026; Yang et al., 2026). Hourglass, by contrast, discovers and refines explicit symbolic strategies, yet its strategies remain confined to single tasks requiring precise transformation logic.

What if this process were scaled across diverse problems? A Meta-Hourglass would accumulate strategies over time through repeated search and feedback, learning reusable meta-strategies within a domain---even for tasks that lack precise transformation logic. Such a system might, for instance, acquire strategies from geometry puzzles and, with extra scaling, extend them to algebra or beyond mathematics. Given a novel task, it would deploy those strategies zero-shot. They would not prescribe exact operations, nor would they guarantee correctness, but, like the meteorologist's, they would shift the baseline away from the arbitrary.

Whether such meta-reasoning can be learned remains open, but the possibility is now visible.

\section*{Acknowledgements}

The author thanks Yi-Fei Liu and Ya-Hua Li for their assistance in verifying the ARC workflow software. Computing resources used in the early stages of this project were provided by Peking University.

\section*{References}

\begin{enumerate}[leftmargin=*,itemsep=4pt]
\small
\item Alemi, A. A., Fischer, I., Dillon, J. V., \& Murphy, K. (2017). Deep variational information bottleneck. \textit{International Conference on Learning Representations (ICLR 2017)}. \href{https://arxiv.org/abs/1612.00410}{arXiv:1612.00410}
\item Chen, L., Li, Z., Lyu, K., Peng, B., \& Wu, H. (2026). The information bottleneck of chain-of-thought and how latent CoT overcomes it. \textit{International Conference on Learning Representations (ICLR 2026)}.
\item Chollet, F. (2019). On the measure of intelligence. \textit{arXiv}. \href{https://arxiv.org/abs/1911.01547}{arXiv:1911.01547}
\item Choudhary, M., Srivatsa, K. V. A., Aeron, G., Bhattacharya, A. R., Dinh, D. K. D., Hanif, I. A., Kotova, D., Kochmar, E., \& Choudhury, M. (2025). UNVEILING: What makes linguistics olympiad puzzles tricky for LLMs? \textit{arXiv}. \href{https://arxiv.org/abs/2508.11260}{arXiv:2508.11260}
\item Dahl, M., Magesh, V., Suzgun, M., \& Ho, D. E. (2024). Large legal fictions: Profiling legal hallucinations in large language models. \textit{Journal of Legal Analysis}, 16(1), 64--93. \href{https://arxiv.org/abs/2401.01301}{arXiv:2401.01301}
\item Davidson, T. R., Falorsi, L., De Cao, N., Kipf, T., \& Tomczak, J. M. (2018). Hyperspherical variational auto-encoders. \textit{arXiv preprint arXiv:1804.00891}. \href{https://arxiv.org/abs/1804.00891}{arXiv:1804.00891}
\item DeepSeek-AI. (2025). DeepSeek-R1: Incentivizing reasoning capability in LLMs via reinforcement learning. \textit{arXiv}. \href{https://arxiv.org/abs/2501.12948}{arXiv:2501.12948}
\item Du, Y., Li, S., Torralba, A., Tenenbaum, J. B., \& Mordatch, I. (2023). Improving factuality and reasoning in language models through multiagent debate. \textit{International Conference on Machine Learning (ICML 2023)}. \href{https://arxiv.org/abs/2305.14325}{arXiv:2305.14325}
\item Ellis, K., Wong, C., Nye, M., Sablé-Meyer, M., Cary, L., Morales, L., Hewitt, L., Solar-Lezama, A., \& Tenenbaum, J. B. (2021). DreamCoder: Bootstrapping inductive program synthesis with wake-sleep library learning. \textit{ACM-SIGPLAN Symposium on Programming Language Design and Implementation (PLDI 2021)}. \href{https://doi.org/10.1145/3453483.3454080}{DOI: 10.1145/3453483.3454080}
\item Franzen, D., Disselhoff, J., \& Hartmann, D. (2025). Product of experts with LLMs: Boosting performance on ARC is a matter of perspective. \textit{International Conference on Machine Learning (ICML 2025)}. \href{https://arxiv.org/abs/2505.07859}{arXiv:2505.07859}
\item Fraser-Taliente, K., Kantamneni, S., Ong, E., Mossing, D., Lu, C., Bogdan, P. C., Ameisen, E., Chen, J., Kishylau, D., Pearce, A., Tarng, J., Wu, A., Wu, J., Zhang, Y., Ziegler, D. M., Hubinger, E., Batson, J., Lindsey, J., Zimmerman, S., \& Marks, S. (2026). Natural language autoencoders produce unsupervised explanations of LLM activations. \textit{Transformer Circuits Thread}.
\item Geirhos, R., Rubisch, P., Michaelis, C., Bethge, M., Wichmann, F. A., \& Brendel, W. (2019). ImageNet-trained CNNs are biased towards texture; increasing shape bias improves accuracy and robustness. \textit{International Conference on Learning Representations (ICLR 2019)}. \href{https://arxiv.org/abs/1811.12231}{arXiv:1811.12231}
\item Goyal, S., \& Dan, S. (2025). IOLBench: Benchmarking LLMs on linguistic reasoning. \textit{arXiv}. \href{https://arxiv.org/abs/2501.04249}{arXiv:2501.04249}
\item Gu, Y., Tafjord, O., Kim, H., Moore, J., Le Bras, R., Clark, P., \& Choi, Y. (2024). SimpleToM: Exposing the gap between explicit ToM inference and implicit ToM application in LLMs. \textit{arXiv}. \href{https://arxiv.org/abs/2410.13648}{arXiv:2410.13648}
\item Hong, S., Zhuge, M., Chen, J., Zheng, X., Cheng, Y., Wang, J., Zhang, C., Wang, Z., Yau, S. K. S., Lin, Z., Zhou, L., Ran, C., Xiao, L., Wu, C., \& Schmidhuber, J. (2024). MetaGPT: Meta programming for a multi-agent collaborative framework. \textit{International Conference on Learning Representations (ICLR 2024)}. \href{https://arxiv.org/abs/2308.00352}{arXiv:2308.00352}
\item Huang, J., Chen, X., Mishra, S., Zheng, H. S., Yu, A. W., Chi, E. H., \& Le, Q. V. (2023). Large language models cannot self-correct reasoning yet. \textit{International Conference on Learning Representations (ICLR 2024)}. \href{https://arxiv.org/abs/2310.01798}{arXiv:2310.01798}
\item Kazemi, M., Fatemi, B., Bansal, H., Palowitch, J., Anastasiou, C., Mehta, S. V., Jain, L. K., Aglietti, V., Jindal, D., Chen, P., Dikkala, N., Tyen, G., Liu, X., Shalit, U., Chiappa, S., Olszewska, K., Tay, Y., Tran, V. Q., Le, Q. V., \& Firat, O. (2025). BIG-Bench Extra Hard. \textit{Proceedings of the 63rd Annual Meeting of the Association for Computational Linguistics (ACL 2025)}. \href{https://arxiv.org/abs/2502.19187}{arXiv:2502.19187}
\item Kingma, D. P., \& Welling, M. (2014, April). Auto-encoding variational bayes. In \textit{Proceedings of the International Conference on Learning Representations (ICLR)}. \href{https://arxiv.org/abs/1312.6114}{arXiv:1312.6114}
\item Lei, S., Cheng, Z., Jia, K., \& Tao, D. (2025). Revisiting LLM reasoning via information bottleneck. \textit{arXiv}. \href{https://arxiv.org/abs/2507.18391}{arXiv:2507.18391}
\item Li, X., \& Eisner, J. (2019). Specializing word embeddings (for parsing) by information bottleneck. \textit{Proceedings of the 2019 Conference on Empirical Methods in Natural Language Processing (EMNLP 2019)}. \href{https://arxiv.org/abs/1910.00163}{arXiv:1910.00163}
\item Lian, D.-C., Huang, R.-S., Chen, P.-E., Lim, C., Lin, Y.-K., Tseng, G.-Y., Yang, T.-C., Lin, Z.-Y., Chen, P.-C., \& Hsieh, S.-K. (2025). LingBench++: A linguistically-informed benchmark and reasoning framework for multi-step and cross-cultural inference with LLMs. \textit{arXiv}. \href{https://arxiv.org/abs/2507.16809}{arXiv:2507.16809}
\item Lin, Z.-L., Shih, Y.-F., \& Hsieh, S.-K. (2025). Probing large language models in reasoning and translating complex linguistic puzzles. \textit{arXiv}. \href{https://arxiv.org/abs/2502.00817}{arXiv:2502.00817}
\item Liu, N. F., Lin, K., Hewitt, J., Paranjape, A., Bevilacqua, M., Petroni, F., \& Liang, P. (2024). Lost in the middle: How language models use long contexts. \textit{Transactions of the Association for Computational Linguistics}, 12, 277--294. \href{https://arxiv.org/abs/2307.03172}{arXiv:2307.03172}
\item Liu, Y., Xu, C., Zhou, Y., Li, Z., \& Xu, Q. (2025). DeepRTL: Bridging Verilog understanding and generation with a unified representation model. \textit{International Conference on Learning Representations (ICLR 2025)}. \href{https://arxiv.org/abs/2502.15832}{arXiv:2502.15832}
\item Madaan, A., Tandon, N., Gupta, P., Hallinan, S., Gao, L., Wiegreffe, S., Alon, U., Dziri, N., Prabhumoye, S., Yang, Y., Gupta, S., Majumder, B. P., Hermann, K., Welleck, S., Yazdanbakhsh, A., \& Clark, P. (2023). Self-refine: Iterative refinement with self-feedback. \textit{Advances in Neural Information Processing Systems 36 (NeurIPS 2023)}. \href{https://arxiv.org/abs/2303.17651}{arXiv:2303.17651}
\item Mitchell, M., Palmarini, A. B., \& Moskvichev, A. (2023). Comparing humans, GPT-4, and GPT-4V on abstraction and reasoning tasks. \textit{arXiv}. \href{https://arxiv.org/abs/2311.09247}{arXiv:2311.09247}
\item Moskvichev, A., Odouard, V. V., \& Mitchell, M. (2023). The ConceptARC benchmark: Evaluating cognitive capabilities of language models. \textit{Transactions on Machine Learning Research}.
\item OpenAI. (2024). Learning to reason with LLMs. \href{https://openai.com/index/learning-to-reason-with-llms/}{OpenAI Blog}.
\item Qian, C., Liu, W., Liu, H., Chen, N., Dang, Y., Li, J., Yang, C., Chen, W., Su, Y., Cong, X., Xu, J., Li, D., Liu, Z., \& Sun, M. (2024). ChatDev: Communicative agents for software development. \textit{Annual Meeting of the Association for Computational Linguistics (ACL 2024)}. \href{https://arxiv.org/abs/2307.07924}{arXiv:2307.07924}
\item Sanz-Guerrero, M., \& Von Der Wense, K. (2025). Corrective in-context learning: Evaluating self-correction in large language models. \textit{Workshop on Insights from Negative Results in NLP (co-located with NAACL 2025)}. \href{https://arxiv.org/abs/2503.16022}{arXiv:2503.16022}
\item Shi, F., Chen, J., Misra, K., Scales, N., Dohan, D., Chi, E. H., Schärli, N., \& Zhou, D. (2023). Large language models can be easily distracted by irrelevant context. \textit{International Conference on Machine Learning (ICML 2023)}. \href{https://arxiv.org/abs/2302.00093}{arXiv:2302.00093}
\item Snell, C., Lee, J., Xu, K., \& Kumar, A. (2024). Scaling LLM test-time compute optimally can be more effective than scaling model parameters. \textit{arXiv}. \href{https://arxiv.org/abs/2408.03314}{arXiv:2408.03314}
\item Suzgun, M., Yuksekgonul, M., Bianchi, F., Jurafsky, D., \& Zou, J. (2026). Dynamic Cheatsheet: Test-time learning with adaptive memory. \textit{Proceedings of the 19th Conference of the European Chapter of the Association for Computational Linguistics (EACL 2026)}. \href{https://arxiv.org/abs/2504.07952}{arXiv:2504.07952}
\item Tishby, N., Pereira, F. C., \& Bialek, W. (1999). The information bottleneck method. \textit{Proceedings of the 37th Annual Allerton Conference on Communication, Control, and Computing}. \href{https://arxiv.org/abs/physics/0004057}{arXiv:physics/0004057}
\item Tishby, N., \& Zaslavsky, N. (2015). Deep learning and the information bottleneck principle. \textit{IEEE Information Theory Workshop}. \href{https://arxiv.org/abs/1503.02406}{arXiv:1503.02406}
\item Tsui, K. (2025). Self-correction bench: Uncovering and addressing the self-correction blind spot in large language models. \textit{arXiv}. \href{https://arxiv.org/abs/2507.02778}{arXiv:2507.02778}
\item Vincent, P., Larochelle, H., Bengio, Y., \& Manzagol, P. A. (2008, July). Extracting and composing robust features with denoising autoencoders. In \textit{Proceedings of the 25th international conference on Machine learning} (pp. 1096--1103). \href{https://doi.org/10.1145/1390156.1390294}{DOI: 10.1145/1390156.1390294}
\item Wang, X., Wei, J., Schuurmans, D., Le, Q., Chi, E., Narang, S., Chowdhery, A., \& Zhou, D. (2023). Self-consistency improves chain of thought reasoning in language models. \textit{International Conference on Learning Representations (ICLR 2023)}.
\item Wen, X., Liu, Z., Zheng, S., Xu, Z., Ye, S., Wu, Z., Liang, X., Wang, Y., Li, J., Miao, Z., Bian, J., \& Yang, M. (2026). Reinforcement learning with verifiable rewards implicitly incentivizes correct reasoning in base LLMs. \textit{International Conference on Learning Representations (ICLR 2026)}. \href{https://arxiv.org/abs/2506.14245}{arXiv:2506.14245}
\item West, P., Holtzman, A., Buys, J., \& Choi, Y. (2019). BottleSum: Unsupervised and self-supervised sentence summarization using the information bottleneck principle. \textit{Proceedings of the 2019 Conference on Empirical Methods in Natural Language Processing (EMNLP 2019)}. \href{https://arxiv.org/abs/1909.07405}{arXiv:1909.07405}
\item Wong, C., Ellis, K., Tenenbaum, J. B., \& Andreas, J. (2021). Leveraging language to learn program abstractions and search heuristics. \textit{International Conference on Machine Learning (ICML 2021)}. \href{https://arxiv.org/abs/2106.11053}{arXiv:2106.11053}
\item Yang, M., Piao, J., Xia, X., Lan, X., Chen, J., Gong, Y., \& Li, Y. (2026). SkillMaster: Toward autonomous skill mastery in LLM agents. \textit{arXiv}. \href{https://arxiv.org/abs/2605.08693}{arXiv:2605.08693}
\item Yin, Z., Sun, Q., Guo, Q., Wu, J., Qiu, X., \& Huang, X. (2023). Do large language models know what they don't know? \textit{Findings of the Association for Computational Linguistics: ACL 2023}. \href{https://arxiv.org/abs/2305.18153}{arXiv:2305.18153}
\item Yu, Z., Zhou, C., Lin, Y., Zhang, H., Ye, H., Cui, J., Pan, Z., Zhao, J., \& Ding, Y. (2026). ChipBench: A next-step benchmark for evaluating LLM performance in AI-aided chip design. \textit{arXiv}. \href{https://arxiv.org/abs/2601.21448}{arXiv:2601.21448}
\item Zhang, Q., Wang, D., Qian, H., Li, Y., Zhang, T., Huang, M., Xu, K., Li, H., Yan, L., \& Qiu, H. (2025). Understanding the dark side of LLMs' intrinsic self-correction. \textit{Proceedings of the 63rd Annual Meeting of the Association for Computational Linguistics (ACL 2025)}. \href{https://arxiv.org/abs/2412.14959}{arXiv:2412.14959}
\end{enumerate}

\newpage
\appendix

\section{Alignment with ARC-AGI-2 Competition Rules (Competition Mode)}

The official ARC-AGI-2 leaderboard evaluates submissions under a strict \textbf{Pass@2} metric: for each hidden test query, the system may submit exactly two predicted pixel grids, without access to test ground truth at any point during test-time computation. To align with this protocol, we implement a \textbf{Competition Mode} for Hourglass: independent runs of the full pipeline (Induction $\to$ Deduction $\to$ Execution $\to$ Refinement) are executed up to a maximum of $K$ times, and two candidates are selected via a consensus-and-ranking mechanism. \textbf{Slot 1} is filled by the first prediction reproduced by $N$ independent runs (the \textit{consistency threshold}; relaxed to $N=2$ if no consensus emerges within 25 runs), declaring early stopping. \textbf{Slot 2} is filled by a dedicated Ranker agent---an auxiliary LLM call that scores all non-consensus candidates by support-set performance, the declared symbolic rules $(\phi, T)$, and internal reasoning consistency, selecting the top-ranked alternative. If no consensus is reached at all, the Ranker selects the two most promising candidates from the full pool. Due to resource constraints, we tested only the standard Hourglass configuration in Competition Mode, though certain ablation variants may outperform it on individual metrics.

To characterize the compute--accuracy trade-off, we swept the consensus threshold ($N=2,3,4$, with $K=32,60,60$ respectively) on the 120 public evaluation puzzles using Gemini 3.1 Pro. Results, evaluated with official partial-credit scoring, are summarized in Table~\ref{tab:compmode}.

\begin{table}[htbp]
\centering
\small
\begin{tabular}{@{}lccc@{}}
\toprule
\textbf{Metric} & \textbf{C-2} & \textbf{C-3} & \textbf{C-4} \\
\midrule
Max Runs & 32 & 60 & 60 \\
Slot 1 Acc. & 85.8\% & 86.1\% & 86.5\% \\
Pass@2 Acc. & 87.8\% & 87.8\% & 86.9\% \\
Oracle Ceiling & 91.4\% & 94.3\% & 95.4\% \\
Avg. Cost (USD) & $\sim$\$3.98 & $\sim$\$8.83 & $\sim$\$11.30 \\
\bottomrule
\end{tabular}
\caption{Competition Mode evaluation and inference cost (ARC-AGI-2, Gemini 3.1 Pro). Slot 1 Acc. = consensus prediction accuracy; Pass@2 Acc. = Slot 1 $\cup$ Slot 2 accuracy; Oracle Ceiling = any single run correct.}
\label{tab:compmode}
\end{table}

The $N=2$ configuration offers the best cost--accuracy balance; raising the threshold to $N=3$ more than doubles cost with no accuracy gain, and $N=4$ increases cost further while accuracy slightly declines. This decline reflects a saturation effect: the Ranker is left choosing among an increasingly low-quality pool of non-consensus candidates as the sampling budget is exhausted pursuing a stricter consensus.

\textbf{Comparison to contemporary solvers.} To contextualize the \$3.98/task, 87.8\% operating point, we compare it against contemporary test-time compute solvers on the ARC-AGI-2 leaderboard (as of mid-2026): Symbolica's Agentica Framework reaches 85.28\% at \$6.94/task using Claude Opus 4.6; Imbue's Code Evolution Solver reaches 95.1\% at \$8.71/task using extensive evolutionary mutations on Gemini 3.1 Pro; and Google's Gemini 3 Deep Think reaches 84.6\% at \$13.62/task using extended reasoning tokens. Hourglass ($N=2$) offers a competitive cost-performance trade-off for budget-constrained scenarios relative to these systems, achieving comparable accuracy to the more expensive baselines at a lower cost. We emphasize that this is an exploratory secondary result; our primary contribution is the reasoning workflow, not the competition-mode engineering.

We note two caveats relevant to interpreting all numbers in this appendix. First, evaluation is conducted on the \textbf{public} evaluation set; possible pretraining exposure to this set is a shared limitation across all methods compared here, including the leaderboard baselines above, and absolute accuracy may not transfer to the private leaderboard. Second, due to the substantial cost of exhaustive sampling (exceeding \$1,300 for a single full run at $N=4$), each configuration was evaluated with a single run rather than repeated sampling across seeds; the reported numbers are therefore point estimates rather than averages with confidence intervals.

\section{Code Structure Diagnostics}

We noticed that Hourglass-generated code tended to be more modular. To quantify this, we measured AST properties across all submitted solutions.

\begin{table}[htbp]
\centering
\small
\begin{tabular}{@{}llccc@{}}
\toprule
\textbf{Model} & \textbf{Method} & \textbf{Helper functions} & \textbf{Longest function (lines)} & \textbf{Dictionary literals} \\
\midrule
\multirow[t]{3}{*}{GPT-5.5} & Hourglass & 8.4 & 70.7 & 1.4 \\
& Self-Refine & 0.2 & 153.7 & 0.9 \\
& Struct-SR & 1.5 & 147.7 & 0.8 \\
\midrule
\multirow[t]{3}{*}{Gemini 3.1 Pro} & Hourglass & 0.3 & 119.0 & 0.8 \\
& Self-Refine & 0.1 & 121.4 & 0.5 \\
& Struct-SR & 0.1 & 95.9 & 0.5 \\
\bottomrule
\end{tabular}
\caption{Structural code metrics by model and method (mean values).}
\label{tab:codestruct}
\end{table}

On GPT-5.5, Hourglass induces substantially more modular code; on Gemini, structural differences are negligible, yet accuracy still improves markedly (86.7\% vs. 76.9\% best-of-5). Thus, code structure variation is not the primary driver---rule abstraction is.

\section{BBEH-Linguini Implementation Details}

This appendix details the data preprocessing pipeline (C.1) and the design of the text-based Inferer agent (C.2) used for BBEH-Linguini.

\subsection{Data Preprocessing}

The original BBEH-Linguini problems present few-shot demonstrations in heterogeneous, unstandardized formats. To enable automated execution feedback---which requires exact-string comparison of predicted outputs against ground-truth answers on the training pairs---the input and output of every training pair must be programmatically extractable. We therefore standardized all problems into a uniform JSON schema.

We first used \textbf{DeepSeek V4 Pro} to segment each problem into a structured JSON object containing the problem stem, the list of training pairs, and the translation direction. The direction is implicitly encoded by assigning source sentences to the \texttt{input} field and target sentences to the \texttt{output} field; tasks that share the same stem may appear with opposite directions (e.g., low-resource language $\to$ English vs. English $\to$ low-resource language), which require distinct rules.

The original benchmark contains 48 problems in which the few-shot items are not aligned as explicit input--output pairs. Instead, they present two unordered lists---one of sentences in language A and one in language B---and the solver must first infer the correspondence between them before any rule induction can begin. Because our pipeline relies on predefined input--output pairs at training time, these 48 problems were removed.

To ensure the fidelity of the extracted JSON, we conducted a multi-stage verification.

\begin{enumerate}
\item \textbf{Exact-string cross-check}: every string in the JSON was verified to exist verbatim in the original problem text, ensuring no addition or distortion.
\item \textbf{Triple segmentation and cross-validation}: the same DeepSeek V4 Pro was used for a second independent segmentation, and Gemini 3.1 Pro performed a third. The three resulting JSON versions were submitted to DeepSeek V4 Pro for a consistency review. Any discrepancy was flagged.
\item \textbf{Manual audit}: human annotators inspected the segmented outputs for anomalies. Eight problems exhibited segmentation ambiguities that made it impossible to determine the correct label assignment (i.e., which language direction was intended). These were discarded.
\end{enumerate}

Starting from the full set of 200 problems, we removed 48 unordered-pair problems and 8 ambiguous problems, yielding a clean dataset of \textbf{144 problems}. All three methods---Raw Prompt, Self-Refine, and Hourglass---receive identical JSON inputs built from this set.

\subsection{Inferer Agent}

In ARC-AGI-2 and ChipBench, execution feedback is obtained by running generated code. Because BBEH-Linguini requires purely textual rules, we implemented a dedicated \textbf{Inferer} agent to ``execute'' textual rules by generating predicted outputs, analogous to a code interpreter.

The Inferer receives the induced rule together with a set of training examples as few-shot prompts. To prevent the Inferer from bypassing the rule and relying solely on surface-level pattern matching, we adopt a held-out validation protocol.

\begin{itemize}
\item For each run, two training pairs are randomly selected and their outputs are \textbf{withheld}. The Inferer is asked to produce predictions for these two withheld pairs (serving as validation items) and for one test item, all within the same inference pass.
\item If the predictions on both validation items match the ground-truth outputs exactly, the rule is considered to have been applied correctly on the seen distribution. To guard against accidental success, a second round is immediately triggered: two different training pairs are randomly withheld, and the Inferer again predicts both validation items and the test item. If the second round also yields exact matches on the validation pairs, the test prediction from the \textbf{second} round is taken as the final output. The test prediction from the first round is discarded.
\item If any validation pair fails in either round, the run is marked as unsuccessful and contributes to the denominator (but not the numerator) of the pass@k metric.
\end{itemize}

This two-round design ensures that the final output derives from a single inference pass (the second round) without self-consistency voting or multiple sampling, keeping the evaluation protocol aligned with the Raw Prompt baseline. It also provides a stringent correctness check: the rule must generalize across different subsets of the training data before any test output is accepted.

The same Inferer is used for both Hourglass and Self-Refine, ensuring that any performance differences originate entirely from the quality of the induced rules rather than from the execution mechanism.

\section{Prompt Design}

This appendix documents the prompt structure shared across methods, the differences between Hourglass and the Self-Refine baseline, and the rationale behind those differences. Every prompt consists of three components: (i) a \textbf{task description}, which specifies the input format, output requirements, and evaluation protocol; (ii) \textbf{meta-instructions}, which provide auxiliary guidance and warnings about common errors; and (iii) \textbf{output format constraints}, which prescribe how the model should structure its response.

\subsection{Shared Framework and Key Differences}

Both Hourglass and Self-Refine receive identical task descriptions for a given benchmark, ensuring that any performance difference cannot be attributed to asymmetric information about the problem itself. The two methods also share the same set of general meta-instructions that are unrelated to output formatting (e.g., ``do not copy absolute coordinates,'' ``verify consistency across support examples''). The crucial divergence lies in the remaining two components:

\begin{itemize}
\item \textbf{Meta-instructions specific to structured output.} Hourglass modules receive additional instructions that govern the content and form of the symbolic intermediates $\phi$ and $T$ (e.g., ``$\phi$ must use relative positions,'' ``$T$ must refer only to entity types defined in $\phi$''). Self-Refine, which does not produce structured symbolic intermediates, omits these instructions.
\item \textbf{Output format constraints.} Hourglass mandates explicit, labeled fields for $\phi$, $z$, and $T$, encoded as structured natural-language templates. Self-Refine imposes no analogous format constraints; the model is free to organize its reasoning and code in any structure, language, or logical flow it deems appropriate.
\end{itemize}

This design reflects a deliberate trade-off: Hourglass invests in a rigid symbolic interface to enforce the structured bottleneck, while Self-Refine retains maximal flexibility, mimicking the most common deployment of iterative refinement in practice.

\subsection{Meta-Instructions}

The meta-instructions used in all prompts were developed through an iterative process of manual error analysis and LLM-assisted review of failure cases. For \textbf{ARC-AGI-2}, the meta-instructions were refined using outputs from \textbf{Gemini 3.0 Flash} and \textbf{DeepSeek V3.2}, two models that are distinct from---and, in the case of DeepSeek, belong to a different family than---the evaluation models GPT-5.5 and Gemini 3.1 Pro. This separation eliminates the risk of circular optimization where the same model used for evaluation would inform the prompt design. For \textbf{ChipBench} and \textbf{BBEH-Linguini}, the meta-instructions were developed with the assistance of \textbf{Gemini 3.1 Pro}, which is also one of the evaluation backbones. We acknowledge this as a potential source of mild bias; however, its impact is demonstrably limited, as discussed below.

The meta-instructions can be partitioned into two categories:

\begin{itemize}
\item \textbf{General meta-instructions (shared).} These capture task-agnostic pitfalls observed during development---e.g., anchoring on spurious coordinates, overfitting to example indices, conflating support-set order with logical priority. Both Hourglass and Self-Refine receive these instructions in all modules.
\item \textbf{Structural meta-instructions (Hourglass-only).} These govern the production of $\phi$ and $T$ and are tailored to the structured bottleneck---e.g., requirements for origin-independence, entity-based relational descriptions, and modular decomposition of the transformation rule. Because Self-Refine does not generate explicit symbolic intermediates, it does not receive these prompts.
\end{itemize}

\subsection{Output Format Constraints}

Hourglass's Induction and Deduction modules are required to produce outputs conforming to a fixed schema, with explicit fields corresponding to the schema $\phi$, the transient scaffold $z$, and the transformation rule $T$. The concrete field names and internal structure vary by benchmark---e.g., on ARC-AGI-2, $\phi$ is realized as a \textit{Cross-Matrix Invariant Rules} block and $T$ as a \textit{Cue-Guided Transformation Rule} block (\S D.5 gives a plain-language account of the full template)---but the underlying constraint is the same across all three benchmarks: each field is delimited by explicit markers and internally organized into labeled subsections, forcing the model to commit to a specific symbolic abstraction before any code or text artifact is produced. This is not a formatting convention; it operationalizes the structured bottleneck itself.

The Self-Refine baseline deliberately omits any output format constraints, on all three benchmarks. The model is instructed to produce a solution artifact (code or text) and may, at its discretion, include explanatory comments, intermediate reasoning, or rule summaries---but none of these are required or parsed by the execution pipeline. This design choice preserves the baseline's generality: imposing arbitrary format constraints on Self-Refine could artificially depress its performance, while the unconstrained variant better reflects standard practice.

\subsection{Design Limitations}

The meta-instructions and output templates were developed without deep domain expertise in electronic design automation or formal linguistics, and with limited engineering resources. Consequently, we cannot claim that the prompts used in this study are optimal for any of the three benchmarks. However, this limitation also means that the reported results partially reflect a realistic, non-expert deployment scenario, strengthening the practical relevance of the findings.

As noted above, the ChipBench and BBEH meta-instructions were iterated with Gemini 3.1 Pro, which also serves as one of the evaluation models. This introduces a theoretical risk of prompt overfitting. However, ablation experiments (\S 5) confirm that the performance gains are driven by the workflow topology rather than by the precise wording of prompts, suggesting that the results are robust to the specific choice of prompts.

\subsection{Prompts for ARC-AGI-2 Hourglass}

This section provides the complete prompts used for the four-stage Hourglass pipeline on ARC-AGI-2.
The prompts are reproduced below with their original structure, but adapted to \LaTeX{} formatting.
Placeholders for dynamic content (e.g., \texttt{[STEP1\_FULL\_ANALYSIS]}) are shown as they appear during actual API calls.

\subsubsection{Step 1: Induction}

\textbf{Task Description}

You are an image puzzle analysis expert. Analyze 2D numerical matrices where numbers (0-9) represent pixel colors. Perform analysis in \textbf{three parts}:

\begin{enumerate}
  \item \textbf{Part 1}: Analyze \textbf{all input matrices} — identify common characteristics and per-matrix object features.
  \item \textbf{Part 2}: Analyze \textbf{all output matrices} — identify common characteristics and per-matrix object features.
  \item \textbf{Part 3}: Consolidate \textbf{Cross-Matrix Invariant Rules} separately for inputs and outputs — the abstract, reusable rules distilled from Parts 1-2. This summary will be passed to downstream steps; write it to be self-contained and actionable without needing the per-matrix details.
\end{enumerate}

\textbf{Important}: The list may include \textbf{test input} matrices; analyze them with the input group in Part 1 (they have no expected output).

\textbf{Core Principles}
\begin{enumerate}
  \item \textbf{Macro-Level from Micro-Level}: Macro-Level description must be \textbf{grounded in cross-matrix object commonality}. First extract objects per matrix (Micro-Level); then summarize \textbf{recurring object types, shared structural roles, and common patterns} across matrices — this forms the stabilized Macro-Level view.
  \item \textbf{Functional Abstraction}: Do not just identify colors/coordinates. Define objects by their \textbf{Roles} (e.g., ``Container'', ``Anchor'', ``Path'', ``Palette'', ``Symmetry Axis'', ``Frame''). Use topological terms like ``Lanes'', ``Strips'', ``Closed Holes'', and ``Boundaries''.
  \item \textbf{Micro-Level (Object Extraction)}: For each matrix, extract \textbf{all} objects. Categorize into \textbf{Salient Objects} and \textbf{Other Objects}. \textbf{Coverage}: every non-zero pixel belongs to exactly one object.
  \item \textbf{Topology Awareness}: Explicitly distinguish between 4-connectivity and 8-connectivity. Identify ``Global Background'' vs ``Internal Cavities''. Note if an object acts as a ``Ruler'' (central/median base) or a ``Container'' (locking internal pixels).
  \item \textbf{Relativization}: Describe positions using \textbf{relative logic} (e.g., ``distance to nearest edge'', ``parity of row index'', ``offset from Anchor object'') rather than absolute coordinates like (5,5) which may fail to generalize.
  \item \textbf{Complex objects}: Use \textbf{multi-aspect} (shape, color, connectivity, symmetry) and \textbf{hierarchical} (whole → parts → sub-parts; coarse → fine) description.
\end{enumerate}

\textbf{Important Guidelines}
\begin{itemize}
  \item \textbf{Container Recognition}: If a closed loop exists, prioritize defining its interior as a work region.
  \item \textbf{Axis/Median Rule}: Identify if objects are centered on a ``Median Line'' or ``Axis''.
  \item \textbf{Parity Awareness}: Note if object behavior changes based on odd/even dimensions or positions.
  \item \textbf{No Fabrication}: If no shared patterns exist, leave themes empty.
  \item \textbf{Consistency}: Use the same functional labels (e.g., ``Anchor'') across all pairs.
\end{itemize}

\paragraph{Output Format}

\subparagraph{Part 1: All Input Matrices}

\textbf{Common Characteristics (Input)}:
\begin{itemize}
  \item \textbf{has\_similarities}: [Yes or No]
  \item \textbf{shared\_structural\_themes}: [Recurring object roles, functional types, common patterns. Note topological invariants like connectivity type.]
\end{itemize}

\textbf{Individual Input Matrices}:
\textbf{matrix\_1} (Input):
\begin{itemize}
  \item \textbf{structural\_description}: [Macro-level description using shared functional vocabulary.]
  \item \textbf{combination\_strategy}: [How elements/objects combine (e.g., layering, nesting, tiling).]
  \item \textbf{Object List}:
  \begin{itemize}
    \item \textbf{Salient Objects}: Salient Object 1 (Role/Type, Rough Description, Relative Positional Description), ...
    \item \textbf{Other Objects}: Other Object 1 (Role/Type, Description, Positional Description), ...
  \end{itemize}
  \item \textbf{object\_distribution}: [Spatial layout, relative grouping, and repetition patterns.]
\end{itemize}

\textbf{matrix\_2} (Input): [\dots]
\textit{Repeat for each input matrix (including test input).}

\subparagraph{Part 2: All Output Matrices}

\textbf{Common Characteristics (Output)}:
\begin{itemize}
  \item \textbf{has\_similarities}: [Yes or No]
  \item \textbf{shared\_structural\_themes}: [Recurring roles and patterns across all outputs.]
\end{itemize}

\textbf{Individual Output Matrices}:
\textbf{matrix\_1} (Output):
\begin{itemize}
  \item \textbf{structural\_description}: [Macro-level description.]
  \item \textbf{combination\_strategy}: [How objects combine.]
  \item \textbf{Object List}:
  \begin{itemize}
    \item \textbf{Salient Objects}: ...
    \item \textbf{Other Objects}: ...
  \end{itemize}
  \item \textbf{object\_distribution}: [Spatial layout, relative grouping.]
\end{itemize}

\textbf{matrix\_2} (Output): [\dots]

\subparagraph{Part 3: Cross-Matrix Invariant Rules}

Consolidate the abstract rules from Parts 1-2 into a self-contained reference. Downstream steps will read this without access to the per-matrix details above. Separate input and output rules clearly.

\textbf{Input Invariant Rules}:
\begin{verbatim}
## Grid Paradigm (Input)
- **Grid Dimensions**: [Fixed HxW / variable / H==W / other pattern]
- **Background Color**: [color index 0-9]
- **Connectivity**: [4-connectivity / 8-connectivity / mixed]

## Object Classes (cross-input)
- **Role [Name]**: [Invariant properties: topology, color range, shape constraints,
  spatial constraints, relative positioning rules — derived from Common
  Characteristics in Part 1]

## Layout & Composition Rules (Input)
- [Layering rules, intersection handling, relative spacing, alignment patterns
  — derived from Part 1]
\end{verbatim}

\textbf{Output Invariant Rules}:
\begin{verbatim}
## Grid Paradigm (Output)
- **Grid Dimensions**: [How output dimensions relate to input dimensions]
- **Background Color**: [color index 0-9]
- **Connectivity**: [4-connectivity / 8-connectivity / mixed]

## Object Classes (cross-output)
- **Role [Name]**: [Invariant properties: topology, color range, shape constraints,
  spatial constraints, relative positioning rules — derived from Common
  Characteristics in Part 2]

## Layout & Composition Rules (Output)
- [Layering rules, intersection handling, relative spacing, alignment patterns
  — derived from Part 2]
\end{verbatim}

\bigskip
\hrule
\bigskip
\texttt{[INSERT SOURCE PUZZLE MATRICES HERE]}

\subsubsection{Step 2: Deduction}

\textbf{Task Description}

You are an image puzzle analysis expert. Given the cross-matrix invariant rules and training pairs, derive a unified \textbf{Transformation Rule} that maps input objects to output objects across all pairs.

The transformation must:
\begin{itemize}
  \item Operate on the abstract roles and layouts defined in the invariant rules (e.g., ``modify the Anchor object'', not ``move the blue pixels'').
  \item Make every operation dependent on an origin-independent \textbf{generation cue} extracted from the input data (e.g., ``color of the top-left-most pixel'', ``count of objects matching Role X'').
  \item Include explicit \textbf{tie-break rules} for any ambiguous selection or ordering.
  \item Use parametric logic (H, W, offsets, //, \%) — avoid hardcoded dimensions or coordinates.
\end{itemize}

\textbf{Meta-Guidance from Cross-Puzzle Analysis}:
Successful transformations avoid absolute values and embrace \textbf{Parametric Logic}. They use mathematical operators like \texttt{//} (floor div) and \texttt{\%} (modulo) for periodicity and relative offsets (e.g., \texttt{(limit - 1) - x}) for symmetry. They define explicit \textbf{Tie-break rules} (e.g., ``if multiple candidates, pick min\_r then min\_c'') to ensure determinism.

\textbf{Core Principles}
\begin{enumerate}
  \item \textbf{Holistic Parametrization}: Analyze all pairs to find \textbf{Geometric Constants} (e.g., ``Output is always 2x input size'', ``Center is always at height // 2'').
  \item \textbf{Cue Binding}: Every operation MUST have a \textbf{generation cue} that is origin-independent (e.g., ``color of the top-left-most pixel'', ``count of objects with size > 1''). Omit operations with no generalizable cue.
  \item \textbf{Explicit Mapping}: Write transformation functions as f(r, c) where possible. Account for non-square matrices (handle height vs width explicitly).
  \item \textbf{Topology Conservation}: Ensure logic preserves connectivity (e.g., a ``Path'' object in input should map to a continuous structure in output).
  \item \textbf{Tie-break Explicitly}: If an operation selects an object, define the sort order (e.g., ``smallest area'', ``top-most row'').
\end{enumerate}

\textbf{Important Guidelines}
\begin{itemize}
  \item \textbf{Avoid Early Concretization}: Do not assume fixed sizes (like 7x7) if the training data shows variation. Use variables (H, W).
  \item \textbf{Constant Offset Detection}: Be precise about \texttt{size - 1} vs \texttt{size} when calculating boundaries.
  \item \textbf{Parity-Awareness}: If logic involves a ``center'', specify behavior for even-sized dimensions (e.g., ``pick the smaller of the two center indices'').
  \item \textbf{Functional Consistency}: Ensure the assigned ``Role'' (e.g., ``Container'') is respected by the ``Generation Method''.
\end{itemize}

\paragraph{Output Format}

\subparagraph{Transformation Rule}
\begin{verbatim}
## Output Grid Dimensions
- [Formula relating output dimensions to input dimensions]

## Cue-Guided Operations

### Operation Group 1
- **Transformed Object**: [Role from the invariant rules]
- **Transformation Goal**: [High-level visual outcome]
- **Generation Method**: [Parametric description of the operation]
- **Generation Cue**: [The dynamic value or attribute that triggers/drives
  this operation]
- **Cue Extraction Method**: [Precise rule to extract the cue from raw input
  — must work for all pairs]

### Operation Group 2
- [...]

## Global Tie-Break & Ambiguity Resolution
- [Deterministic rules to resolve spatial edge-cases, overlaps, or selection
  order (e.g., "if multiple candidates, pick min_r then min_c")]
\end{verbatim}

\bigskip
\hrule
\bigskip

\texttt{[STEP1\_FULL\_ANALYSIS]}

\texttt{[FORMATTED\_TRAINING\_PAIRS]}

\subsubsection{Step 3: Implementation}

\textbf{Task Description}

You are an image puzzle transformation expert. Given the invariant rules, transformation rule, training pairs, and test input, produce:

\begin{enumerate}
  \item \textbf{Part 1}: A concrete execution plan (natural language), including any hardcoded constants with justification.
  \item \textbf{Part 2}: A complete, runnable Python script implementing the plan.
\end{enumerate}

\textbf{Core Principles}
\begin{enumerate}
  \item \textbf{Deterministic Logic}: All choices must be derived from input data. Use deterministic tie-breaks (e.g., \texttt{sorted(objs, key=lambda x: (x.r\_min, x.c\_min))}).
  \item \textbf{Constant +1/-1 Precision}: When handling boundaries or mirroring, carefully implement \texttt{(limit - 1) - x} to avoid off-by-one errors.
  \item \textbf{Topology Check}: For path-following or flooding, ensure the pixel count/connectivity is conserved. Use 4-neigh vs 8-neigh consistently.
  \item \textbf{Helper Robustness}: Helpers must \textbf{raise ValueError} on ambiguity or missing cues.
  \item \textbf{Pythonic Initialization}: Avoid shallow copy issues. Use \texttt{new\_grid = [[bg\_color for \_ in range(W)] for \_ in range(H)]}.
\end{enumerate}

\textbf{Hardcoding Policy}
\begin{itemize}
  \item \textbf{Allowed}: (1) Training-derived structural constants (e.g., ``background is always color 0''). (2) Cross-puzzle patterns (Checkerboards, Frames, Borders). (3) Test-input-specific properties (H, W).
  \item \textbf{Not allowed}: Direct mapping from training inputs to outputs. No hardcoded pixel coordinates from training data. No color-based special cases (e.g., \texttt{if color == 3}).
  \item \textbf{Part 1 Requirement}: Explicitly state \textbf{What} is hardcoded, \textbf{Why}, and \textbf{How}.
\end{itemize}

\textbf{Script Robustness (Meta-Guidance)}
\begin{itemize}
  \item \textbf{Axis/Median Rule}: When a ``central line'' is needed, derive it as \texttt{(dim - 1) // 2} or similar, documenting how even dimensions are handled.
  \item \textbf{Peeling Strategy}: If the transformation involves complex backgrounds, consider ``peeling'' (masking) salient objects, performing global operations, then ``pasting'' them back.
  \item \textbf{Parity Awareness}: Check if \texttt{r \% 2 == 0} is required for checkerboards or alternating patterns.
\end{itemize}

\textbf{Important Guidelines}
\begin{itemize}
  \item \textbf{1:1 Topology}: If expanding a path, ensure no unintended gaps or overlaps occur.
  \item \textbf{Origin-Independence}: Use logic that works regardless of whether the ``Anchor'' is at (0,0) or (H-1, W-1).
  \item \textbf{No Process Exit}: Do not use \texttt{sys.exit()}.
\end{itemize}

\paragraph{Output Format}

\subparagraph{Part 1: Execution Plan}
[Step-by-step algorithm: Initialization → Object Extraction → Cue Detection → Transformation → Output Assembly. Include a \textbf{Hardcoded Constants} subsection listing any training-derived constants (e.g., background color, connectivity type) with justification for each.]

\subparagraph{Part 2: Python Script}
\begin{verbatim}
[Python code only.]
\end{verbatim}

\textbf{Script Output Contract}
\begin{itemize}
  \item Output exactly one Python fenced code block.
  \item Must define \texttt{transform\_puzzle(input\_grid: list[list[int]]) -> list[list[int]]}.
  \item Return \texttt{List[List[int]]} (not numpy array).
  \item Use only standard Python libraries. Use \texttt{collections.deque} for BFS/connected components.
  \item Helpers must raise ValueError on invalid or ambiguous states.
  \item No hardcoded pixel coordinates from training data. No color-based special cases (e.g., \texttt{if color == 3}). Derive all structural parameters from the input data.
  \item No placeholder code, TODO text, \texttt{sys.exit()}, or prose inside the code block.
\end{itemize}

\bigskip
\hrule
\bigskip

\texttt{[INVARIANT\_RULES]}

\texttt{[TRANSFORMATION\_RULE]}

\texttt{[FORMATTED\_MATRICES]}

\subsubsection{Step 4: Refinement}

\textbf{Task Description}

You are an expert puzzle analyst. Revise the program so it correctly maps every training input to its expected output.

\textbf{Meta-Guidance for Revision}:
\begin{enumerate}
  \item \textbf{Root-Cause Diagnosis}: Do not just describe \emph{where} pixels differ. Diagnose \emph{why} the logic failed (e.g., ``The mirror axis was assumed to be at W//2 but should be at W//2 - 1'').
  \item \textbf{Surgical Fix (One-Inch Rule)}: Successful fixes usually involve changing a single constant or mapping (e.g., \texttt{r} to \texttt{c}). If you are adding massive \texttt{if-else} blocks, your geometric model is likely wrong.
  \item \textbf{Anti-Patching}: Avoid \texttt{if color == 3: ...}. Instead, find the parametric reason why color 3 behaves differently.
  \item \textbf{Parity-Awareness}: Check if the failure occurs only on even/odd dimensions.
  \item \textbf{Anti-Regression}: If your fix involves ``simply counting colors'' while losing ``spatial position'', stop. This is a sign of model degradation.
\end{enumerate}

\textbf{Important Guidelines}
\begin{itemize}
  \item \textbf{Avoid Threshold Oscillation}: Do not just tweak \texttt{if count > 4} to \texttt{> 5}. Re-evaluate the object definition.
  \item \textbf{Topology Conservation}: If the fix involves movement, ensure objects don't merge or disappear unless intended.
  \item \textbf{Parity-Awareness}: If dimensions changed from training to test, check if your division/centering logic is robust.
\end{itemize}

\paragraph{Output Format}

\subparagraph{Part 1: Diagnosis}
\begin{itemize}
  \item \textbf{Error comparison}: [Actual vs Expected — where and how predictions deviate.]
  \item \textbf{Root cause}: [The logic error — e.g., off-by-one in symmetry axis, wrong connectivity mode, missing parity handling.]
  \item \textbf{Surgical fix justification}: [Why this correction fixes the root cause for ALL pairs, not just the failing one.]
\end{itemize}

\subparagraph{Part 2: Revised Invariant Rules}
[Full updated invariant rules.]

\subparagraph{Part 3: Revised Transformation Rule}
[Full updated transformation rule.]

\subparagraph{Part 4: Revised Execution Plan}
[Full updated plan matching the revised rules.]

\subparagraph{Part 5: Revised Python Script}
\begin{verbatim}
[Python code only.]
\end{verbatim}

\textbf{Script Output Contract}
\begin{itemize}
  \item Output exactly one Python fenced code block in Part 5.
  \item Must define \texttt{transform\_puzzle(input\_grid: list[list[int]]) -> list[list[int]]}.
  \item Return \texttt{List[List[int]]} (not numpy array).
  \item Use only standard Python libraries. Use \texttt{collections.deque} for BFS/connected components.
  \item Helpers must raise ValueError on invalid or ambiguous states.
  \item No hardcoded pixel coordinates from training data. No color-based special cases.
  \item No placeholder code, TODO text, \texttt{sys.exit()}, or prose inside the code block.
\end{itemize}

\bigskip
\hrule
\bigskip

\texttt{[INVARIANT\_RULES]}

\texttt{[TRANSFORMATION\_RULE]}

\texttt{[CURRENT\_CODE]}

\texttt{[ERROR\_TRACES\_JSON]}

\end{document}